\documentclass[runningheads]{llncs}
\usepackage{parskip}
\usepackage{graphicx}
\usepackage{verbatim}
\usepackage{booktabs}
\usepackage[numbers]{natbib}

\begin{document}
\title{ORION Grounded in Context: Retrieval-Based Method for Hallucination Detection}
%
%
\author{
    Assaf Gerner\inst{1} \and
    Netta Madvil\inst{1} \and
    Nadav Barak\inst{1} \and
    Alex Zaikman\inst{1} \and
    Jonatan Liberman\inst{1} \and
    Liron Hamra\inst{1} \and
    Rotem Brazilay\inst{1} \and
    Shay Tsadok\inst{1} \and
    Yaron Friedman\inst{1} \and
    Neal Harow\inst{1} \and
    Noam Bressler\inst{1} \and
    Shir Chorev\inst{1} \and
    Philip Tannor\inst{1}
}
%
%
\institute{
    Deepchecks, Ramat Gan, Israel
}
\maketitle              
\pagestyle{plain}
\begin{abstract}
Despite advancements in grounded content generation, production Large Language Models (LLMs) based applications still suffer from hallucinated answers. We present "Grounded in Context" - a member of Deepchecks' ORION (Output Reasoning-based InspectiON) family of lightweight evaluation models. It is our framework for hallucination detection, designed for production-scale long-context data and tailored to diverse use cases, including summarization, data extraction, and RAG. Inspired by RAG architecture, our method integrates retrieval and Natural Language Inference (NLI) models to predict factual consistency between premises and hypotheses using an encoder-based model with only a 512-token context window. Our framework identifies unsupported claims with an F1 score of 0.83 in RAGTruth's response-level classification task, matching methods that trained on the dataset, and outperforming all comparable frameworks using similar-sized models.
\end{abstract}

\section{Introduction}
In natural language generation tasks such as Retrieval-Augmented Generation (RAG) and abstractive summarization, hallucinations—instances where generated text contains contradictory or fabricated information in comparison to a reference text—persist as a significant challenge in practical applications, despite considerable advancements in grounded content generation \cite{Ji_2023}. In RAG systems, these hallucinations emerge from inconsistencies between retrieved data and generated content, substantially undermining output reliability. Similarly, in abstractive summarization, models may produce information not directly inferable from source text, resulting in summaries that introduce spurious details or contradict the original document.

At Deepchecks, our primary objective is to evaluate Large Language Model (LLM) based applications with particular emphasis on detecting and mitigating such hallucinations. In practical scenarios, users retrieve data fragments from diverse sources that are frequently unstructured or "noisy". Furthermore, contemporary LLMs can process extensive contexts, necessitating evaluation frameworks capable of handling substantial data volumes both efficiently and precisely.

To address these challenges, we present ORION (Output Reasoning-based InspectiON) - a family of lightweight evaluation models designed to assess the quality of LLM and SLM outputs. It performs structured, multistep analysis across dimensions such as factual consistency, information density, task relevance, and avoidance. ORION enables precise, reliable, and explainable evaluation pipelines critical for quality assurance, compliance, and iterative development of language-based systems.

A crucial member of the ORION family is "Grounded in Context", a RAG-inspired methodology for hallucination detection. Our approach decomposes outputs into factual statements and retrieves a dedicated context for each statement. Statement-context pairs are evaluated for factual consistency, with individual scores aggregated into a comprehensive metric. Our methodology demonstrates superior performance compared to models of comparable size when evaluated on the RAGTruth dataset \cite{niu2024ragtruthhallucinationcorpusdeveloping}, despite the exclusion of its training corpus from our method's training data.

\section{Related Work}
Research in hallucination detection has grown significantly in recent years, mainly driven by the increasing deployment of RAG systems. Detection methods generally focus on identifying factual inaccuracies by comparing generated content with retrieved evidence, highlighting discrepancies between them.

Some approaches have incorporated transformer-based models trained for Natural Language Inference (NLI) to evaluate consistency between generated claims and external data. Such models can be encoder or decoder-based. NLI models categorize hypotheses as entailed, neutral, or contradictory in relation to a given premise. By establishing entailment relationships while disregarding the distinction between neutrality and contradiction \cite{wu2023wecheckstrongfactualconsistency}\cite{gekhman2023trueteacherlearningfactualconsistency}, NLI-based methodologies provide a structured framework to assess the substantiation of generated text by source data.

Nevertheless, processing extensive documents within the constrained context window of NLI models presents a significant practical limitation. Several methodological solutions have emerged to address this challenge, ranging from computing token-level entailment scores for individual context documents and subsequently aggregating these values \cite{belyi2024lunaevaluationfoundationmodel}, to leveraging advancements in encoder-based models to accommodate expanded context windows \cite{kovács2025lettucedetecthallucinationdetectionframework}.

\section{Method}
\subsection{Problem Statement}
To explain our architectural approach, we need to address several key challenges in designing an effective hallucination detection system:
\begin{enumerate}
\item \textbf{Non-factual statements:} LLMs often generate many non-factual statements to make content more readable, such as titles and greetings. These statements aren't information-dense enough to count as hallucinations, but they typically don't align factually with the retrieved context.
\item \textbf{Long context:} Leading encoder-based models for classification tasks had relatively limited context windows until recent developments \cite{warner2024smarterbetterfasterlonger}. Even now, contexts retrieved in RAG systems tend to exceed what encoder-based models can process.
\item \textbf{Prediction resolution:} Current methods vary in how finely they detect hallucinations—whether at the token level, proposition level, or across the entire sequence.
\end{enumerate}
\subsection{The Grounded in Context Solution}
\label{sec:solution}

Grounded in Context takes inspiration from RAG by addressing the long context problem through creating specific premises for each claim in the output, pulling the most relevant sections from the document collection. This represents a straightforward approach when implementing proposition-level analysis.

We focus on proposition-level analysis because we assume that models can learn the task more easily since entailment is fundamentally a relationship between propositions, while still offering enough explainability for root cause analysis. However, it’s worth noting that token-level classification offers better explainability. While proposition-level methods can highlight hallucinated statements, token-level approaches can pinpoint exactly which parts of a statement contribute to hallucinations.

\begin{figure}[ht]
\centering
\includegraphics[width=0.6\textwidth]{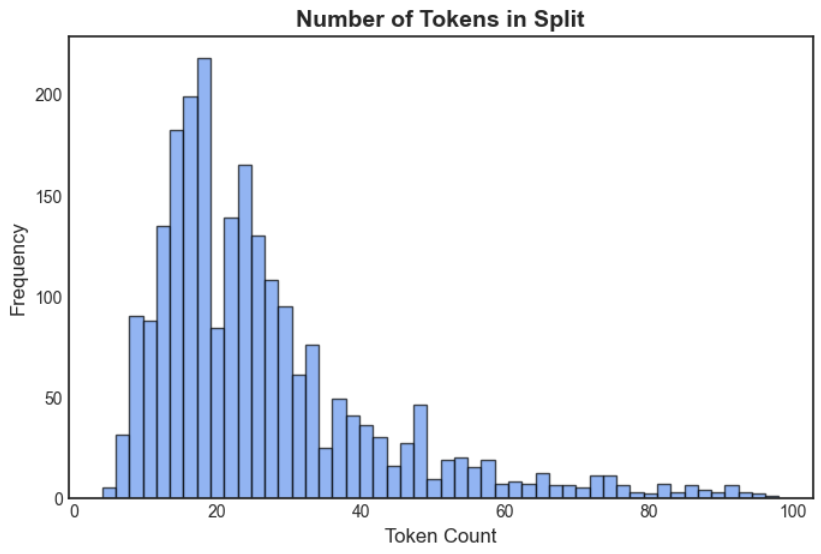}
\caption{: Distribution of the number of tokens in a claim after LLM generated sequences were split by our chunker without setting a maximum token size.}
\label{fig:output-chunks-tokens}
\end{figure}

Our method consists of the following steps:
\begin{enumerate}
\item Split the output $O$ into claims $C = {c_1, c_2, ..., c_n}$ using our recursive text chunker $\mathcal{T}$ with parameters maximum chunk size $s_{max}$ and maximum chunk overlap $o_{max}$. Our experiments indicate that setting $s_{max} = 60$ tokens captures almost all of cases without unnaturally segmenting sentences, as Figure \ref{fig:output-chunks-tokens} shows.
\item Filter out non-factual claims from $C$ using a compact encoder-based factual claims classifier $\mathcal{F}$, resulting in filtered set $C' \subseteq C$.
\item Split the context $D$ into chunks $D = {d_1, d_2, ..., d_m}$ using the same chunker $\mathcal{T}$. The chunking parameters are calibrated to fit a number of chunks that is chosen dynamically depending on the claim length within a 512-token context window after accounting for claim token length.
\item For each claim $c_i \in C'$, retrieve the (dynamically chosen) $k$ most relevant chunks $R_i = {d_{i1}, d_{i2}, ..., d_{ik}} \subset D$.
\item Score each claim-chunk pair $(c_i, d_{ij})$ using the entailment probability $p_{ent}(c_i, d_{ij})$ from an NLI model $\mathcal{M}$.
\item Aggregate the entailment scores using function $\mathcal{A}$ that applies greater weighting to negative classifications, effectively penalizing hallucinations.
\end{enumerate}

\section{Evaluation}
We initially assessed the method on our private evaluation datasets, which more closely simulate production-like use cases, where it demonstrated strong performance. In addition, we benchmark our method on the RAGTruth dataset \cite{niu2024ragtruthhallucinationcorpusdeveloping}, a popular word-level hallucination detection dataset which chooses 3 recognized tasks for response generation: Question Answering, Data-to-text Writing, and News Summarization.

Benchmarking the method exactly as described above with \cite{li2024angleoptimizedtextembeddings} used for retrieval and \cite{wu2023wecheckstrongfactualconsistency} used as an NLI model yields an F1 score of 0.72, surpassing Luna, which was the state-of-the-art until recently.

\begin{table*}[ht]
\centering
\small
\label{tab:performance}
\resizebox{\textwidth}{!}{%
\begin{tabular}{lcccccccccccc}
\toprule
& \multicolumn{3}{c}{\textbf{QUESTION ANSWERING}} & \multicolumn{3}{c}{\textbf{DATA-TO-TEXT WRITING}} & \multicolumn{3}{c}{\textbf{SUMMARIZATION}} & \multicolumn{3}{c}{\textbf{OVERALL}} \\
\cmidrule(lr){2-4} \cmidrule(lr){5-7} \cmidrule(lr){8-10} \cmidrule(lr){11-13}
\textbf{Method} & \textbf{Prec.} & \textbf{Rec.} & \textbf{F1} & \textbf{Prec.} & \textbf{Rec.} & \textbf{F1} & \textbf{Prec.} & \textbf{Rec.} & \textbf{F1} & \textbf{Prec.} & \textbf{Rec.} & \textbf{F1} \\
\midrule
\multicolumn{13}{c}{\textbf{Trained on the Dataset}} \\
\midrule
Finetuned Llama-2-13B & 61.6 & 76.3 & 68.2 & 85.4 & 91.0 & 88.1 & 64.0 & 54.9 & 59.1 & 76.9 & 80.7 & 78.7 \\
RAG-HAT & 76.5 & 73.1 & 74.8 & 92.9 & 90.3 & \textbf{91.6} & 77.7 & 59.8 & 67.6 & 87.3 & 80.8 & \textbf{83.9} \\
Luna & 37.8 & 80.0 & 51.3 & 64.9 & 91.2 & 75.9 & 40.0 & 76.5 & 52.5 & 52.7 & 86.1 & 65.4 \\
lettucedetect-large-v1 & 65.9 & 75.0 & 70.2 & 90.4 & 86.7 & 88.5 & 64.0 & 55.9 & 59.7 & 80.4 & 78.05 & 79.2 \\
\midrule
\multicolumn{13}{c}{\textbf{Not Trained on the Dataset}} \\
\midrule
Prompt\textsubscript{gpt-4-turbo} & 33.2 & 90.6 & 45.6 & 64.3 & 100.0 & 78.3 & 31.5 & 97.6 & 47.6 & 46.9 & 97.9 & 63.4 \\
SelCheckGPT\textsubscript{gpt-3.5-turbo} & 35.0 & 58.0 & 43.7 & 68.2 & 82.8 & 74.8 & 31.1 & 56.5 & 40.1 & 49.7 & 71.9 & 58.8 \\
LMvLM\textsubscript{gpt-4-turbo} & 18.7 & 76.9 & 30.1 & 68.0 & 76.7 & 72.1 & 23.2 & 81.9 & 36.2 & 36.2 & 77.8 & 49.4 \\
ChainPoll\textsubscript{gpt-3.5-turbo} & 33.5 & 51.3 & 40.5 & 84.6 & 35.1 & 49.6 & 45.8 & 48.0 & 46.9 & 54.8 & 40.6 & 46.7 \\
RAGAS Faithfulness & 31.2 & 41.9 & 35.7 & 79.2 & 50.8 & 61.9 & 64.2 & 29.9 & 40.8 & 62.0 & 44.8 & 52.0 \\
Trulens Groundedness & 22.8 & 92.5 & 36.6 & 66.9 & 96.5 & 79.0 & 40.2 & 50.0 & 44.5 & 46.5 & 85.8 & 60.4 \\
\textbf{Grounded in Context (Ours)} & 88.4 & 91.9 & \textbf{90.1} & 64.4 & 82.2 & 72.2 & 85.35 & 87.1 & \textbf{86.2} & 79.4 & 87.1 & 83.05 \\
\bottomrule
\end{tabular}
}
\caption{Performance comparison across the various tasks in RAGTruth. We compare our results with several approaches presented in Luna \cite{belyi2024lunaevaluationfoundationmodel}, RAGTruth \cite{niu2024ragtruthhallucinationcorpusdeveloping}, RAG-HAT \cite{song-etal-2024-rag} and LettuceDetect \cite{kovács2025lettucedetecthallucinationdetectionframework}.}
\end{table*}

Deepchecks' current method builds upon the principles in Section \ref{sec:solution} and incorporates additional improvements in chunking, retrieval, and context construction, \textbf{alongside a proprietary NLI model.} It achieves an F1 score of 0.83, second only to RAG-HAT (0.84), a substantially larger model that was trained on RAGTruth, unlike our model, for which the dataset remains out-of-distribution.

It is notable that our framework's improvements were less pronounced in the data-to-text assignment, which we attribute to a combination of challenges in effectively chunking formatted data and the limited representation of such data in our training set, rather than an inherent deficiency in the methodology itself. As our training data regarding structured text primarily consists of feature extraction use-cases where entities are extracted in isolation, we hypothesize that these specific use-cases explain the low-precision, high-recall ratio observed in the data-to-text task.

\section{Conclusion}
In this paper, we presented Grounded in Context, an approach for hallucination detection that effectively addresses the challenges of long contexts and varying statement types. By decomposing outputs into discrete factual statements and retrieving dedicated context for each, our method demonstrates strong performance on hallucination detection tasks and the best results on RAGTruth for a model of its size even though it wasn't trained on the dataset.

\subsection{Future Work}
As our research predates the release of newer comparable encoder models with enhanced long-context capabilities \cite{warner2024smarterbetterfasterlonger}, we faced constraints in selecting optimal chunking and retrieval parameters. These limitations may lead to suboptimal contextualization of document chunks and necessitate a low retrieval value $k$, reducing the probability of identifying the correct chunks for grounding output claims. Such limitations become particularly pronounced when applying models to real-world, noisy data environments, which present significantly greater challenges than the controlled conditions of curated benchmark datasets.

We hypothesize that integrating our method with models such as ModernBERT, which efficiently processes sequences of up to 8,192 tokens, has the potential to substantially enhance performance by overcoming existing retrieval and contextual limitations. We plan to investigate this prospect in the coming months.

%
%
%
\bibliographystyle{unsrtnat}
\bibliography{grounded_in_context}

\end{document}